\documentclass[letterpaper]{article} 
\usepackage[preprint]{aaai2027}  
\usepackage[hyphens]{url} 
\usepackage{graphicx} 
\usepackage{natbib} 
\usepackage{caption} 
\usepackage{algorithm}
\usepackage{algorithmic}
\usepackage{array}

\usepackage{newfloat}
\usepackage{listings}
\DeclareCaptionStyle{ruled}{labelfont=normalfont,labelsep=colon,strut=off} 
\floatstyle{ruled}
\newfloat{listing}{tb}{lst}{}
\floatname{listing}{Listing}

\usepackage{booktabs}

\title{BLADE: Boundary-Expanded and Layer-Adaptive \\ Dynamic Exit for Efficient LLM Reasoning}

\author{
    Keshu Fu\textsuperscript{\rm 1}\equalcontrib,
    Keqin Peng\textsuperscript{\rm 1}\equalcontrib,
    Jun Bai\textsuperscript{\rm 2}\equalcontrib,
    Shuhan Qin\textsuperscript{\rm 1},
    Chen Li\textsuperscript{\rm 1},
    Junzhu Liang\textsuperscript{\rm 3},
    Yefei Chen\textsuperscript{\rm 4}\\
    Jiaqi Li\textsuperscript{\rm 2},
    Yuanxin Ouyang\textsuperscript{\rm 1}\protect\thanks{Corresponding author.}
}

\affiliations{
    \textsuperscript{\rm 1}School of Computer Science and Engineering, Beihang University\\
    \textsuperscript{\rm 2}Beijing Institute for General Artificial Intelligence, BIGAI\\
    \textsuperscript{\rm 3}Academy for Advanced Interdisciplinary Studies, Peking University
\\
    \textsuperscript{\rm 4}School of Computer Science and Technology, East China Normal University\\
    \textbf{Correspondence:} fukeshu@buaa.edu.cn,
    oyyx@buaa.edu.cn
}

\usepackage{amsmath}
\usepackage{amssymb}
\usepackage{arydshln}

\begin{document}

\maketitle

\begin{abstract}
Large language models often improve task performance by generating long reasoning traces, but the resulting computation is frequently wasted on redundant verification and revision. Existing probe-based early-exit approaches mainly inspect explicit self-doubt expressions, leaving many earlier termination opportunities undetected. Expanding inspection to ordinary reasoning boundaries improves coverage, but also exposes highly diverse intermediate states whose predictive information may reside in different hidden layers. We present Boundary-Expanded and Layer-Adaptive Dynamic Exit for Efficient LLM Reasoning (BLADE), a lightweight framework that dynamically terminates reasoning by estimating whether the generated prefix is sufficient for correct answering. BLADE constructs multi-granular checkpoints from sentence, self-doubt, and paragraph boundaries, and derives robust training labels through repeated answer completions. It further learns a compact subset of informative probe layers instead of relying on fixed choices or expensive representations from all layers. At inference time, calibrated predictions are combined with checkpoint-specific confirmation rules to balance responsiveness and premature-exit risk. 
Experiments on five benchmarks and two Qwen3 reasoning models show that BLADE preserves near-baseline accuracy while reducing generated tokens by 24.8\% on Qwen3-8B and 15.8\% on Qwen3-4B. Ablation studies further confirm the benefits of diverse checkpoints and automatic layer selection, demonstrating an effective approach to more efficient LLM reasoning.
\end{abstract}

\section{Introduction}

Large Language Models (LLMs) achieve strong performance through extended reasoning traces \citep{han2026structured}. Early work showed that Chain-of-Thought (CoT) prompting improves arithmetic, commonsense, and symbolic reasoning \citep{wei2022chain}. Recent studies show that additional inference-time computation can improve reasoning performance \citep{snell2025scaling,wu2025native}.

However, more computation is not always beneficial. LLMs may overthink simple problems, repeatedly verify established conclusions, or continue generating after reaching a correct answer \citep{chen2024donot,sui2025stop,peng2025revisiting}. Such unnecessary reasoning increases inference cost and can degrade answer quality through redundant revision or deviation from a correct solution \citep{akgul2025lynx,yang2025dynamic}.

\begin{figure}[t]
    \centering
    \includegraphics[
        width=\columnwidth
    ]{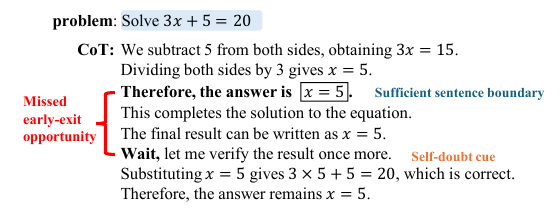}
    \caption{
    A missed early-exit opportunity: the correct answer appears before ``Wait,'' but self-doubt-only monitoring overlooks it and permits redundant verification.
    }
    \label{fig:motivation}
\end{figure}

A practical approach to mitigating overthinking is to enable reasoning early exit with a lightweight probe \citep{zhang2025reasoning,akgul2025lynx}. At selected checkpoints, the probe examines hidden states and predicts whether the current reasoning prefix is sufficient for a correct final answer. This requires training only a small prediction module, without modifying or retraining the base model.
In this way, prior work uses self-doubt markers, such as ``wait,'' ``however,'' and ``let me reconsider,'' as natural exit checkpoints \citep{fu2025reasoning,yang2025dynamic}. These markers often signal a shift from problem solving to reflection or revision, indicating that a viable solution may already have been reached. Probing internal confidence at such positions can therefore identify opportunities to terminate unnecessary reasoning.

However, self-doubt checkpoints miss many sufficient reasoning states. As shown in Fig.~\ref{fig:motivation}, a regular sentence boundary may already complete a key calculation or establish the correct conclusion before any self-doubt checkpoint appears. We therefore expand the checkpoint set to include both self-doubt markers and sentence boundaries, enabling earlier and broader detection of sufficient prefixes.
This broader coverage introduces heterogeneous reasoning states, ranging from incomplete derivations and intermediate conclusions to reflection and revision. Their sufficiency signals may be most informative at different model depths, making fixed layer choices difficult to generalize. Concatenating all layers avoids manual selection but creates a large, redundant probe input with substantial parameter and inference overhead.

To address these issues, we propose \textbf{BLADE}, a \textbf{B}oundary-Expanded and \textbf{L}ayer-\textbf{A}daptive \textbf{D}ynamic \textbf{E}xit framework for efficient LLM reasoning. BLADE formulates early exit as prefix-sufficiency prediction and coordinates three decisions: where to inspect the reasoning trajectory, which hidden layers to probe, and when to terminate generation. Because sparse self-doubt checkpoints can miss earlier sufficient states, \textbf{Multi-Granular Reasoning Checkpoints (MGRC)} augments them with sentence and paragraph boundaries, providing broader coverage at complementary reasoning granularities. 

The resulting checkpoints represent heterogeneous reasoning states, whose sufficiency signals may be distributed differently across model layers. To avoid both brittle manual layer choices and costly all-layer concatenation, \textbf{Adaptive Probe-Layer Selection (APLS)} identifies a compact subset of informative layers for a lightweight sufficiency probe. Finally, because dense sentence checkpoints are less semantically reliable than sparse self-doubt checkpoints, BLADE adopts a calibrated, checkpoint-aware stopping policy: confident self-doubt predictions can trigger immediate exit, whereas sentence-level predictions require consecutive confirmation to prevent premature termination.

Experiments on five mathematical reasoning benchmarks with two Qwen3 backbones \citep{yang2025qwen3} show that BLADE achieves a favorable accuracy--efficiency trade-off among the evaluated early-exit methods. Compared with full CoT generation, BLADE reduces generated tokens by $24.8\%$ on Qwen3-8B and $15.8\%$ on Qwen3-4B while largely preserving accuracy. Further analyses validate both adaptive designs. As shown in Supplementary Table~3, on MATH-500\citep{hendrycks2021measuring,lightman2024verify} with Qwen3-8B, combining sentence and self-doubt checkpoints improves accuracy from $84.89\%$ to $85.56\%$ while reducing generated tokens by $17.8\%$ compared with self-doubt checkpoints alone. At approximately $86\%$ accuracy, APLS uses $17.6\%$ fewer tokens than random four-layer selection, demonstrating the importance of both broad checkpoint coverage and informative layer selection.

Our main contributions are as follows:
\begin{itemize}
\item We reveal a fundamental trade-off in checkpoint design: self-doubt-only probing provides limited coverage, whereas broader checkpoint coverage introduces heterogeneous reasoning states.

\item We propose BLADE, which combines multi-granular reasoning checkpoints, adaptive probe-layer selection, and checkpoint-aware stopping to enable reliable and lightweight dynamic reasoning exit.

\item Experiments on multiple benchmarks and backbones demonstrate favorable accuracy-efficiency trade-offs, reducing generated tokens by up to $24.8\%$ while largely preserving accuracy. Further analyses validate the benefits of checkpoint diversity and automatic layer selection.

\end{itemize}

\begin{figure*}[t]
    \centering
    \includegraphics[
        width=\textwidth,
    ]{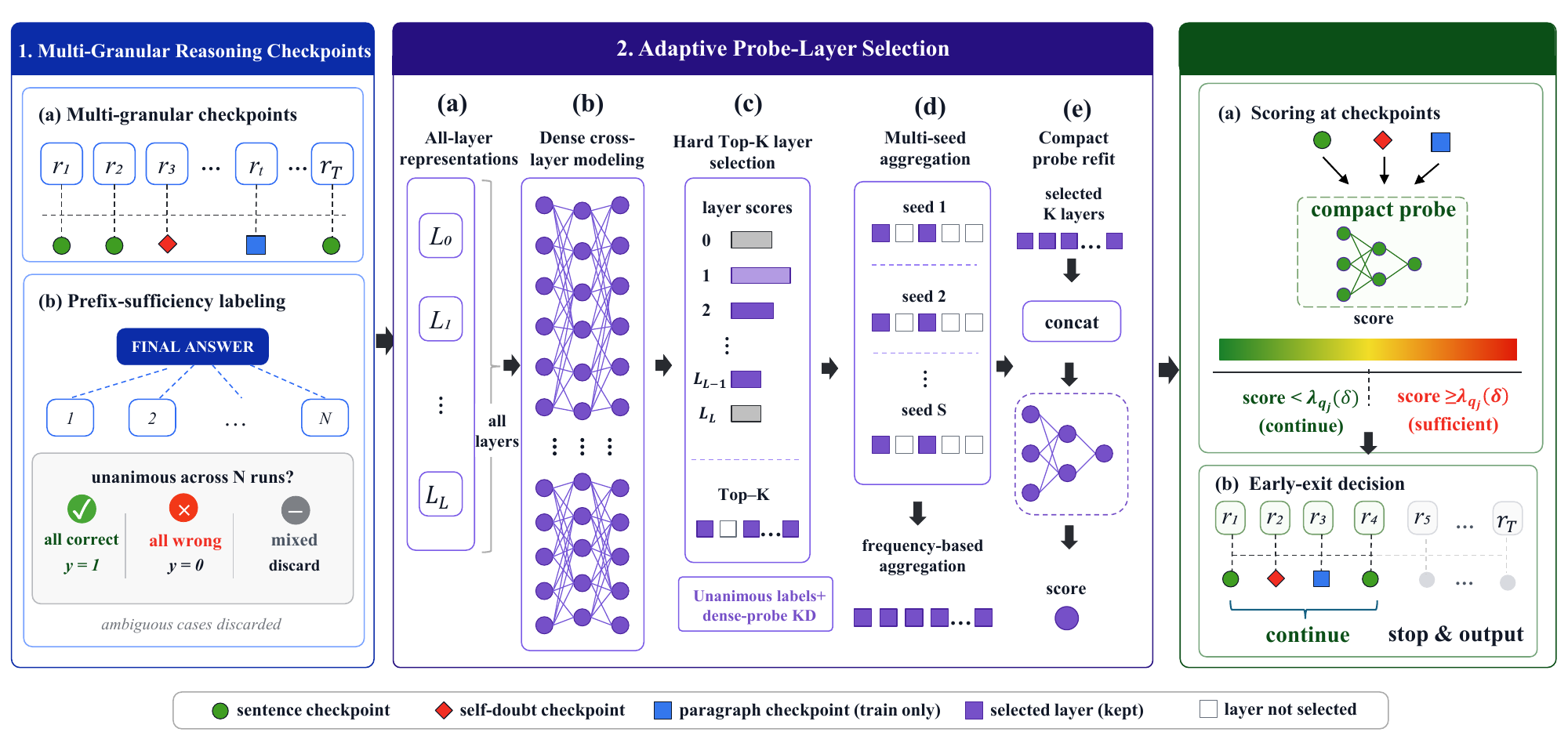}
\caption{
Overview of BLADE. MGRC constructs multi-granular checkpoints and
retains unanimous forced-completion outcomes as prefix-sufficiency
labels. APLS freezes the dense cross-layer modeling, learns hard Top-$K$ masks
across multiple runs, aggregates the selected subsets, and refits a
compact probe. At inference, the calibrated probe applies
checkpoint-type-aware temporal stopping.
}
    \label{fig:framework}
\end{figure*}
\section{Related Work}

\paragraph{Overthinking in Long-CoT Reasoning.}
Long-CoT reasoning can improve performance through additional test-time computation, but may continue after sufficient information for a correct answer has already been obtained. Prior work analyzes this overthinking behavior from outcome- and process-level perspectives \citep{chen2024donot,sui2025stop}. In particular, \citet{peng2025revisiting} identify self-doubt-driven repeated verification as a major source of redundancy and mitigate it through prompting. However, self-doubt does not mark every sufficient reasoning state. BLADE therefore treats it as one signal among multiple checkpoints and performs online prefix-sufficiency prediction rather than globally shortening reasoning.

\paragraph{Dynamic Early Exit for Efficient LLM Reasoning.}
Existing methods differ in where they inspect reasoning and how they decide to stop. DEER, LYNX, and DTSR monitor reflection-related positions, such as \texttt{Wait} and other self-doubt cues, using trial answers, hidden-state probes, or prompted sufficiency judgments to trigger early exit \citep{yang2025dynamic,akgul2025lynx,xiang2026thinking}. Reasoning Path Deviation Monitoring instead tracks high-entropy token transitions to detect departures from productive reasoning paths \citep{guan2026mitigating}, while AdaQR performs query-level routing between lightweight and deeper reasoning modes \citep{zhang2025your}. These approaches are often restricted to sparse reflection cues, local token statistics, or coarse query-level decisions. BLADE combines self-doubt and sentence boundaries with checkpoint-aware calibration and temporal confirmation, broadening checkpoint coverage while controlling premature exits.

\paragraph{Prefix-Sufficiency Probing.}
The hidden states of a reasoning model encode signals predictive of answer correctness, enabling self-verification and early exit \citep{zhang2025reasoning}. However, probe effectiveness depends on the layers used: task-relevant information is distributed unevenly across model depth and may span multiple layers \citep{peters2018deep,tenney2019bert,devries2020special}. Using all layers is costly, while fixed subsets may omit informative representations. BLADE addresses this trade-off with APLS, which learns a compact, adaptive layer subset through dense cross-layer modeling and uses only the selected representations at inference.

\section{Methodology}

We formulate reasoning early exit as prefix-sufficiency prediction: determining whether an intermediate reasoning prefix already supports a reliable correct answer.
As shown in Fig.~\ref{fig:framework}, BLADE consists of two training components and a checkpoint-aware inference policy. MGRC constructs diverse candidate states with low-noise sufficiency labels, while APLS selects a compact set of hidden layers for efficient prediction. At inference, the resulting probe applies checkpoint-specific stopping rules to terminate unnecessary reasoning.

\subsection{Problem Formulation}
\label{sec:formulation}

Given a problem $x_i$, a reasoning model generates a token sequence
\begin{equation}
r_i=(r_{i,1},\ldots,r_{i,T_i}).
\end{equation}
At a candidate boundary $t$, the current reasoning prefix is denoted by
$r_{i,\leq t}$, and its layer-wise hidden states are
\begin{equation}
H_{i,t}
=
\left(
h_{i,t,0},\ldots,h_{i,t,L-1}
\right),
\qquad
h_{i,t,\ell}\in\mathbb{R}^{d}.
\end{equation}

We define a binary prefix-sufficiency variable
$Y_{i,t}\in\{0,1\}$, where $Y_{i,t}=1$ indicates that the current prefix
can reliably support a correct final answer. A lightweight probe
estimates
\begin{equation}
p_\theta(i,t)
=
P_\theta(Y_{i,t}=1\mid H_{i,t})
=
\sigma\!\left(
f_\theta\!\left(\Phi(H_{i,t})\right)
\right),
\end{equation}
where $\Phi$ extracts representations from the compact layer subset
selected by APLS.

\subsection{Multi-Granular Checkpoint Supervision}
\label{sec:mgrc}
\label{sec:early-exit}

\paragraph{Multi-granular checkpoint construction.}
MGRC constructs training candidates from three complementary types of
reasoning checkpoints:
\begin{equation}
\mathcal{C}^{\mathrm{train}}_i
=
\mathcal{C}^{\mathrm{sent}}_i
\cup
\mathcal{C}^{\mathrm{doubt}}_i
\cup
\mathcal{C}^{\mathrm{para}}_i.
\end{equation}
Sentence checkpoints broadly cover completed calculations and
intermediate conclusions, self-doubt checkpoints capture reflection and
verification states, and paragraph checkpoints provide additional
coarse-grained supervision. Their combination extends the coverage of
potentially sufficient states beyond sparse explicit self-doubt checkpoints.

\paragraph{Low-noise prefix-sufficiency supervision.}
For each candidate boundary
$t\in\mathcal{C}^{\mathrm{train}}_i$, we force the model to terminate
its current reasoning trajectory and generate an answer $N=16$ times.
Let
\begin{equation}
b_{i,t,k}
=
\mathbb{I}
\left[
\operatorname{Check}
\left(
\hat{a}_{i,t,k},a_i^\star
\right)
=1
\right],
\qquad
k=1,\ldots,N,
\end{equation}
where $\hat{a}_{i,t,k}$ is the answer produced by the $k$-th forced
completion and $a_i^\star$ is the ground-truth answer. We retain only
unanimous outcomes:
\begin{equation}
y_{i,t}
=
\begin{cases}
0, & \sum_{k=1}^{N} b_{i,t,k}=0,\\
1, & \sum_{k=1}^{N} b_{i,t,k}=N,\\
\bot, & \text{otherwise}.
\end{cases}
\end{equation}
Candidates labeled with $\bot$ are excluded from probe training,
reducing the label noise caused by stochastic completions.

\paragraph{Boundary-adaptive dynamic exit.}

At inference, BLADE evaluates sentence and self-doubt checkpoints, which differ in representation and frequency. Applying a uniform stopping rule can therefore cause premature exits at dense sentence boundaries. BLADE instead exits immediately after an accepted self-doubt checkpoint, but requires two consecutive acceptances at sentence checkpoints. Once triggered, the current prefix is retained and followed by a final-answer completion; otherwise, reasoning continues to its natural end.

\subsection{Adaptive Probe-Layer Selection}
\label{sec:apls}

Concatenating all hidden layers captures rich cross-layer information but incurs substantial redundancy and overhead. As shown in Table~\ref{tab:qwen8b_resource_comparison}, the APLS compact probe reduces parameters by approximately 64\%, peak memory by 85\%, and training time per epoch by 90\% relative to dense cross-layer modeling. APLS achieves this efficiency by selecting a compact, adaptive subset of $K$ layers through dense modeling, hard Top-$K$ selection, and multi-seed aggregation.

\paragraph{Dense cross-layer sufficiency modeling.}
Each layer representation is independently normalized and mapped through
a shared projection:
\begin{equation}
u_{i,t,\ell}
=
\operatorname{GELU}
\left(
W_p\operatorname{LN}_{\ell}(h_{i,t,\ell})+b_p
\right).
\end{equation}
The projected features from all layers are concatenated and passed to a
dense prediction head:
\begin{equation}
z_{i,t}^{T}
=
f_T
\left(
[u_{i,t,0};\ldots;u_{i,t,L-1}]
\right).
\end{equation}
The dense model is trained on the unanimous sufficiency labels using
class-balanced binary cross-entropy. After training, the dense model is frozen, and its output logits are
used to guide the optimization of the layer selector.

\paragraph{Budget-constrained hard Top-$K$ selection.}
After freezing the dense model, each layer is assigned a learnable gate
logit $\alpha_\ell$, which is normalized into a soft gate score:
\begin{equation}
\boldsymbol{\pi}
=
\operatorname{softmax}(\boldsymbol{\alpha}).
\end{equation}
Under the fixed layer budget $K$, the forward pass constructs a hard
binary Top-$K$ mask:
\begin{equation}
m_\ell^{H}
=
\mathbb{I}
\left[
\ell\in
\operatorname{TopK}
\left(
\boldsymbol{\pi},K
\right)
\right].
\end{equation}
Since the discrete Top-$K$ operation is not directly differentiable, we
use a straight-through estimator
\citep{bengio2013estimating}:
\begin{equation}
m_\ell
=
m_\ell^{H}
+
\pi_\ell
-
\operatorname{stopgrad}(\pi_\ell).
\end{equation}
Therefore, exactly $K$ layers are active in the forward pass, while the
backward pass propagates gradients to the gate logits through the
continuous soft gate scores.

A temporary selection head evaluates the masked cross-layer
representation:
\begin{equation}
z_{i,t}^{M}
=
f_M
\left(
[
m_0u_{i,t,0};
\ldots;
m_{L-1}u_{i,t,L-1}
]
\right).
\end{equation}
The gates and selection head are jointly optimized using
class-balanced classification supervision and knowledge distillation
from the frozen dense model. The selection head is used only during
layer search and is discarded afterward.

\paragraph{Stability-aware multi-seed aggregation.}
Due to representation redundancy across Transformer layers
\citep{dalvi2020analyzing}, independent selection runs may identify
different but functionally similar layer subsets. Let
$\mathcal{S}^{(r)}$ denote the Top-$K$ subset obtained from run $r$.
We compute the selection frequency of each layer:
\begin{equation}
f_\ell
=
\frac{1}{R}
\sum_{r=1}^{R}
\mathbb{I}
\left[
\ell\in\mathcal{S}^{(r)}
\right].
\end{equation}
The final adaptive subset consists of the $K$ most frequently
selected layers:
\begin{equation}
\mathcal{S}^{\star}
=
\operatorname{TopK}_{\ell}
\left(
f_\ell,K
\right),
\end{equation}
with mean gate rank used only to break frequency ties. After fixing
$\mathcal{S}^{\star}$, we discard the dense model, gates, and temporary
selection head, and refit a compact probe directly on the concatenated
raw hidden states of the selected layers.

\begin{table*}[t]
\centering
{\small
\setlength{\tabcolsep}{1mm}
\begin{tabular*}{\textwidth}{
@{\extracolsep{\fill}}
l*{13}{c}
@{}
}
\toprule
Method & \multicolumn{2}{c}{GSM8K} & \multicolumn{2}{c}{MATH-500} & \multicolumn{2}{c}{AMC23} & \multicolumn{2}{c}{AIME24} & \multicolumn{2}{c}{AIME25} & \multicolumn{3}{c}{Average} \\
\cmidrule(lr){2-3} \cmidrule(lr){4-5} \cmidrule(lr){6-7} \cmidrule(lr){8-9} \cmidrule(lr){10-11} \cmidrule(lr){12-14}
 & Acc. & \#Tok. & Acc. & \#Tok. & Acc. & \#Tok. & Acc. & \#Tok. & Acc. & \#Tok. & Acc. & \#Tok. & AES \\
\midrule
\multicolumn{14}{c}{\textit{Qwen3-8B}} \\
\midrule
Base & 92.3 & 1705 & 89.8 & 4639 & 88.7 & 7822 & 61.4 & 11956 & 51.9 & 13064 & 76.8 & 7837 & 0.000 \\
\textbf{Ours-Mixed} & 93.7 & 733 & 85.6 & 2620 & 87.5 & 5610 & 60.5 & 9476 & 52.2 & 11055 & 75.2 & 5896 & \textbf{0.213} \\
Ours-Doubt & 93.4 & 830 & 87.8 & 3809 & 88.4 & 5827 & 61.7 & 10092 & 52.5 & 11431 & 76.4 & 6684 & 0.187 \\
\hdashline
LYNX-K16 & 93.8 & 811 & 84.2 & 2507 & 87.7 & 5626 & 61.7 & 10448 & 51.5 & 10982 & 76.2 & 6520 & 0.188 \\
LYNX-K1 & 93.0 & 892 & 87.3 & 3389 & 87.0 & 4961 & 60.8 & 10665 & 52.2 & 11247 & 76.2 & 6650 & 0.163 \\
\midrule
\multicolumn{14}{c}{\textit{Qwen3-4B}} \\
\midrule
Base & 85.3 & 1415 & 92.0 & 4325 & 91.4 & 7205 & 58.3 & 11999 & 52.2 & 13148 & 75.8 & 7618 & 0.000 \\
\textbf{Ours-Mixed} & 92.4 & 700 & 86.9 & 2907 & 88.9 & 4512 & 57.7 & 11382 & 52.5 & 11386 & 75.6 & 6414 & \textbf{0.175} \\
Ours-Doubt & 90.7 & 937 & 88.7 & 3500 & 90.0 & 5122 & 58.0 & 11056 & 52.2 & 12299 & 75.4 & 6757 & 0.112 \\
\hdashline
LYNX-K16 & 90.5 & 975 & 87.6 & 2918 & 89.3 & 5116 & 58.0 & 10415 & 52.2 & 11903 & 75.6 & 6843 & 0.109 \\
LYNX-K1 & 90.9 & 969 & 92.0 & 3863 & 90.0 & 5068 & 58.3 & 11380 & 52.2 & 11662 & 75.9 & 6781 & 0.127 \\

\bottomrule
\end{tabular*}
}
\caption{Main results. Accuracy (\%), average generated tokens,
and AES. Benchmark-specific entries use the highest-AES operating point
in the pre-specified $\delta$ grid; Average follows the protocol
in Sec.~\ref{sec:setup}.}
\label{tab:main_best_aes_qwen_models}

\end{table*}

\section{Experiments}

\subsection{Setup}
\label{sec:setup}

\paragraph{Benchmarks and models.}
We evaluate BLADE on five mathematical reasoning benchmarks—GSM8K-test \citep{cobbe2021training}, MATH-500 \citep{lightman2024verify}, AMC 2023 \citep{aimo2024amc}, AIME 2024 \citep{aime24}, and AIME 2025 \citep{aime25}—using Qwen3-8B and Qwen3-4B \citep{yang2025qwen3}. The suite contains 1,919 questions, split into 192 calibration and 1,727 held-out test examples.

\paragraph{Probe training and inference.}

For each backbone, we train the dense teacher and compact probe on a 6,000-question corpus comprising 2,000 examples each from GSM8K train \citep{cobbe2021training}, the numeric-answer subset of MATH train \citep{hendrycks2021measuring}, and DeepScaleR train \citep{luo2025deepscaler}. Data are split at the question level. Both models are trained for 100 epochs with K16 strict-clean supervision, while checkpoint and layer selection use only the internal validation split. At inference, the resulting probe scores sentence and self-doubt boundaries. Architecture and labeling details appear in Sec.~\ref{sec:apls}.

\paragraph{Calibration, reporting, and metrics.}
For each frozen method, conformal thresholds are estimated on the
calibration split at
$\delta\in\mathcal{D}=\{0.002,0.003,0.005,0.01\}$, where $\delta$
controls the stringency of the stopping threshold: smaller values yield
more conservative early-exit decisions. The resulting thresholds are
applied unchanged to the held-out test split. We report answer accuracy, average generated response tokens,
and Accuracy--Efficiency Score (AES) \citep{luo2025o1pruner}; prompt
tokens are excluded because they are shared by all methods. Let $p$ and
$L$ denote the accuracy and average generated-token count of an evaluated
method, respectively, and let $p_b$ and $L_b$ denote those of the
corresponding Full-CoT baseline. AES is defined as
\begin{equation}
\mathrm{AES}
=
\frac{L_b-L}{L_b}
+
\begin{cases}
3\frac{p-p_b}{p_b}, & p \geq p_b, \\[2pt]
-5\frac{p_b-p}{p_b}, & p < p_b.
\end{cases}
\end{equation}
The first term measures relative token savings, while the second rewards
accuracy improvements and penalizes accuracy degradation. Higher AES
therefore indicates a better accuracy--efficiency trade-off.

For each benchmark, tables report the operating point with the highest AES
over $\mathcal{D}$. The \emph{Average} columns first macro-average metrics
across the five benchmarks at each operating point and then average them
over $\mathcal{D}$; they are therefore not averages of the displayed
benchmark-wise best points.

\begin{figure}[t]
    \centering
    \includegraphics[width=0.95\columnwidth]
    {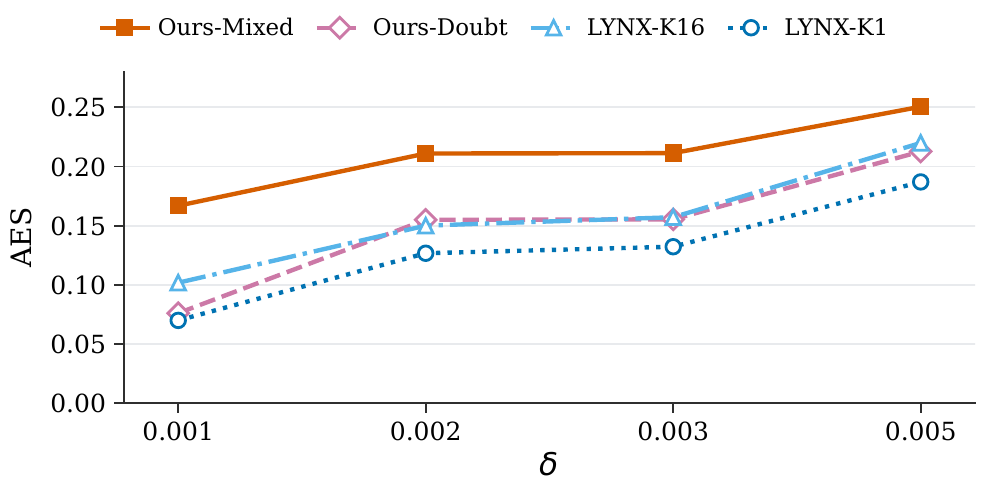}
\caption{Average AES on the five-benchmark Qwen3-8B evaluation suite.
Results are equal-weight macro-averages across the five benchmarks.}
    \label{fig:e1_combined_aes_low_delta}
\end{figure}

\paragraph{Baselines.}
We compare BLADE with Full-CoT and LYNX-style baselines
\citep{akgul2025lynx}. \textsc{LYNX-K1} uses self-doubt checkpoints,
the fixed LYNX layer subset, and forced-exit labels derived from a single
completion. \textsc{LYNX-K16} uses the same checkpoints and fixed layers,
but replaces the single-completion labels with K16 strict-clean
supervision. For layer-selection ablations, we further compare against
final-layer, validation-selected single-layer, fixed four-layer, random
four-layer, adjacent-middle, evenly spaced, and all-layer probes.

\subsection{Main Results}

\paragraph{Accuracy-preserving token savings.}
Table~\ref{tab:main_best_aes_qwen_models} summarizes performance at the
best operating point for each benchmark. On Qwen3-8B, BLADE achieves the
highest average AES ($0.213$), exceeding LYNX-K16 ($0.188$) and LYNX-K1
($0.163$). It reduces the average number of generated tokens from
$7{,}837$ to $5{,}896$, a $24.8\%$ reduction, while retaining $75.2\%$
accuracy compared with $76.8\%$ for Full-CoT. These results show that
BLADE provides a stronger overall balance between answer accuracy and
reasoning cost than the LYNX baselines.

\paragraph{Performance across calibration levels.}
Fig.~\ref{fig:e1_combined_aes_low_delta} reports the combined
five-benchmark results at every $\delta\in\mathcal{D}$. BLADE achieves
the strongest AES throughout the evaluated grid, showing that its
token-saving advantage is not tied to a single calibration setting.
In particular, its advantage remains under stricter stopping thresholds,
where maintaining accuracy is most important.

\paragraph{Generality across models.}
The same pattern holds for Qwen3-4B: BLADE achieves an average AES of
$0.175$, exceeding LYNX-K16 ($0.109$) and LYNX-K1 ($0.127$). It reduces
average generated tokens from $7{,}618$ to $6{,}414$ ($15.8\%$) while
maintaining nearly unchanged average accuracy ($75.6\%$ versus $75.8\%$
for Full-CoT). Together with the Qwen3-8B results, this finding shows that
BLADE remains effective across the two evaluated Qwen3 backbones rather
than relying on the reasoning behavior of a single model.

\begin{table*}[t]
\centering
{\small
\setlength{\tabcolsep}{1mm}
\begin{tabular*}{\textwidth}{
@{\extracolsep{\fill}}
l*{13}{c}
@{}
}
\toprule
Method & \multicolumn{2}{c}{GSM8K} & \multicolumn{2}{c}{MATH-500} & \multicolumn{2}{c}{AMC23} & \multicolumn{2}{c}{AIME24} & \multicolumn{2}{c}{AIME25} & \multicolumn{3}{c}{Average} \\
\cmidrule(lr){2-3} \cmidrule(lr){4-5} \cmidrule(lr){6-7} \cmidrule(lr){8-9} \cmidrule(lr){10-11} \cmidrule(lr){12-14}
 & Acc. & \#Tok. & Acc. & \#Tok. & Acc. & \#Tok. & Acc. & \#Tok. & Acc. & \#Tok. & Acc. & \#Tok. & AES \\
\midrule
\multicolumn{14}{c}{\textit{Qwen3-8B}} \\
\midrule
Base & 92.3 & 1705 & 89.8 & 4639 & 88.7 & 7822 & 61.4 & 11956 & 51.9 & 13064 & 76.8 & 7837 & 0.000 \\
\textbf{Ours: selected K4} & 93.7 & 733 & 85.6 & 2620 & 87.5 & 5610 & 60.5 & 9476 & 52.2 & 11055 & 75.2 & 5896 & \textbf{0.213} \\
\hdashline
Final layer & 92.5 & 682 & 84.0 & 2606 & 83.6 & 5317 & 58.6 & 10091 & 48.8 & 11531 & 71.6 & 5643 & -0.006 \\
LYNX-K4 & 93.3 & 735 & 86.7 & 3344 & 87.7 & 5378 & 60.8 & 9840 & 53.1 & 11141 & 75.1 & 6017 & 0.199 \\
Best single& 92.9 & 712 & 84.9 & 2886 & 87.3 & 5468 & 58.6 & 10269 & 51.9 & 11713 & 73.3 & 5759 & 0.100 \\
Worst single & 92.8 & 991 & 87.6 & 3854 & 85.7 & 6573 & 58.3 & 11397 & 48.1 & 12425 & 71.4 & 6460 & -0.162 \\
Adjacent-middle K4 & 93.0 & 729 & 85.1 & 2776 & 88.9 & 6119 & 61.1 & 11106 & 50.3 & 10598 & 75.0 & 6166 & 0.159 \\
Evenly-spaced K4 & 92.8 & 764 & 84.0 & 2472 & 85.9 & 5243 & 61.4 & 9881 & 52.5 & 11370 & 73.7 & 5755 & 0.151 \\
All layers & 93.8 & 822 & 81.8 & 2220 & 86.6 & 5172 & 60.8 & 10751 & 51.2 & 12047 & 73.7 & 6094 & 0.103 \\
Random K4 (10-seed avg.) & 92.6 & 720 & 86.1 & 3181 & 87.5 & 5691 & 60.6 & 10211 & 51.9 & 11799 & 74.6 & 5945 & 0.166 \\
\midrule
\multicolumn{14}{c}{\textit{Qwen3-4B}} \\
\midrule
Base & 85.3 & 1415 & 92.0 & 4325 & 91.4 & 7205 & 58.3 & 11999 & 52.2 & 13148 & 75.8 & 7618 & 0.000 \\
\textbf{Ours: selected K4} & 92.4 & 700 & 86.9 & 2907 & 88.9 & 4512 & 57.7 & 11382 & 52.5 & 11386 & 75.6 & 6414 & 0.175 \\
\hdashline
Final layer & 92.4 & 907 & 89.1 & 3634 & 90.7 & 6051 & 58.0 & 11777 & 51.2 & 12788 & 75.0 & 6875 & 0.054 \\
LYNX-K4 & 92.2 & 736 & 88.9 & 3141 & 90.7 & 5759 & 57.7 & 11355 & 52.8 & 11605 & 75.5 & 6420 & 0.169 \\
Best single & 92.2 & 886 & 88.9 & 3700 & 90.5 & 5272 & 58.6 & 11931 & 51.9 & 12686 & 75.4 & 6907 & 0.075 \\
Worst single & 89.6 & 779 & 88.9 & 3823 & 90.5 & 6050 & 57.4 & 11766 & 51.2 & 12774 & 74.0 & 6735 & 0.024 \\
Adjacent-middle K4 & 91.7 & 686 & 89.3 & 2971 & 89.3 & 5198 & 57.7 & 10778 & 52.2 & 12053 & 75.0 & 6169 & \textbf{0.176} \\
Evenly-spaced K4 & 92.4 & 738 & 89.6 & 3513 & 89.1 & 4904 & 57.4 & 11251 & 52.5 & 11701 & 76.0 & 6591 & 0.176 \\
All layers & 92.8 & 768 & 90.2 & 3381 & 90.7 & 5978 & 57.1 & 11352 & 51.9 & 11603 & 75.4 & 6476 & 0.147 \\
Random K4 (10-seed avg.) & 92.7 & 755 & 88.7 & 3332 & 89.0 & 4878 & 57.6 & 11573 & 51.4 & 11736 & 75.7 & 6609 & 0.151 \\

\bottomrule
\end{tabular*}
}
\caption{Layer-selection ablation. Accuracy (\%), average
generated tokens, and AES. All results follow the reporting protocol in
Sec.~\ref{sec:setup}; Random K4 is averaged over 10 layer seeds before
operating-point selection.}
\label{tab:ablation_best_aes_qwen_models}
\end{table*}

\subsection{Layer-Selection Ablation}
\label{sec:layer_ablation}

Table~\ref{tab:ablation_best_aes_qwen_models} compares BLADE with single-layer, all-layer, predefined, and random strategies to assess the value of cross-layer modeling and automatic compact layer selection.

\paragraph{Comparison with single-layer probes.}
BLADE substantially outperforms the validation-selected best single-layer
probe, achieving average AES values of $0.213$ versus $0.100$ on
Qwen3-8B and $0.175$ versus $0.075$ on Qwen3-4B. Moreover, even the best single-layer probe underperforms most of the
evaluated four-layer strategies on both backbones. The gap between the best and worst single-layer probes
($0.100$ versus $-0.162$ on Qwen3-8B and $0.075$ versus $0.024$ on
Qwen3-4B) further shows that performance is sensitive to the chosen
layer. Together, these results indicate that a single layer provides an
incomplete representation for prefix-sufficiency prediction, whereas
combining complementary information from multiple depths is beneficial.

\begin{figure*}[t]
    \centering
    \includegraphics[width=\textwidth]{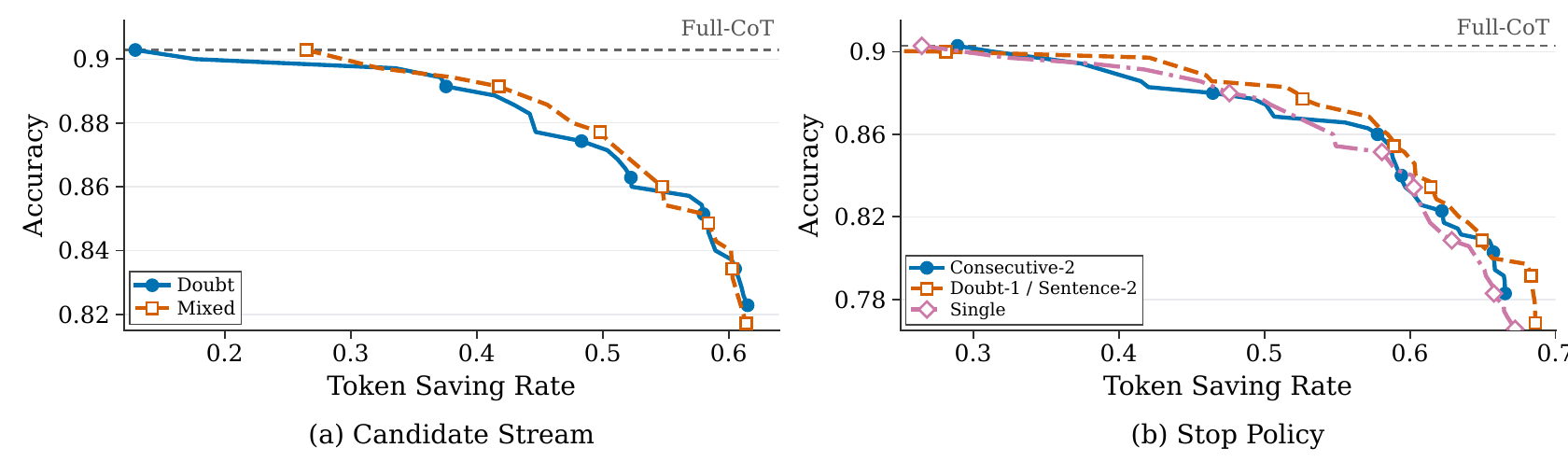}
    \caption{Runtime-policy ablations on MATH-500 with Qwen3-8B.
    All variants use the same frozen probe.
    The left panel shows candidate streams, and the right panel shows stopping policies.}
    \label{fig:e4_runtime_factors}
\end{figure*}

\paragraph{Comparison with all-layer representations.}
Using all layers is not necessarily optimal. BLADE achieves average AES
values of $0.213$ and $0.175$ on Qwen3-8B and Qwen3-4B, respectively,
compared with $0.103$ and $0.147$ for the all-layer probe. Thus, simply
concatenating representations from every layer does not improve
prefix-sufficiency prediction and may introduce redundant or
less task-relevant features that are harder to model effectively. 
As shown in Sec.~\ref{sec:resource_analysis}, retaining dense cross-layer
modeling at deployment incurs substantially greater parameter and
hidden-state access costs. BLADE instead uses APLS to compress dense
cross-layer information into a compact subset that retains strong
early-exit performance.

\paragraph{Comparison with Fixed and Random Layer Selection.}
BLADE avoids reliance on manually specified layer indices. On Qwen3-8B, it achieves the highest average AES among the evaluated strategies ($0.213$), outperforming the fixed LYNX-K4 subset ($0.199$) and random four-layer selection ($0.166$). On Qwen3-4B, BLADE reaches $0.175$, exceeding LYNX-K4 ($0.169$) and random selection ($0.151$), and remaining comparable to adjacent-middle and evenly spaced subsets ($0.176$). Although multiple four-layer combinations can yield similar representations, BLADE consistently identifies a strong compact subset without manual tuning or random search, supporting automatic layer selection in APLS.

\begin{figure}[t]
\centering

\includegraphics[
    width=0.95\columnwidth,
]{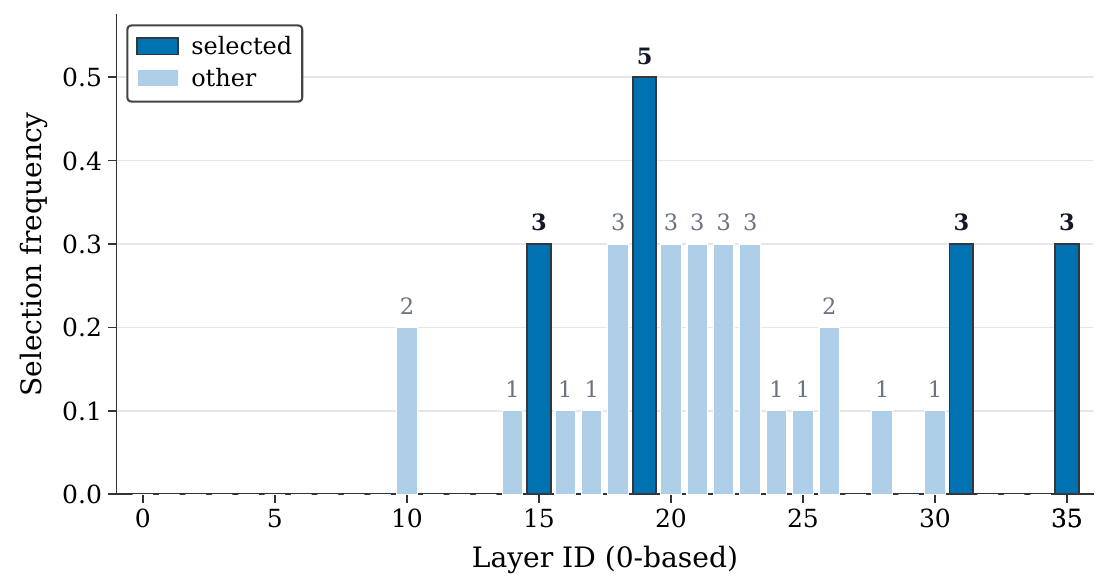}

\vspace{0mm}
{\footnotesize (a) Qwen3-8B\par}

\includegraphics[
    width=0.95\columnwidth,
]{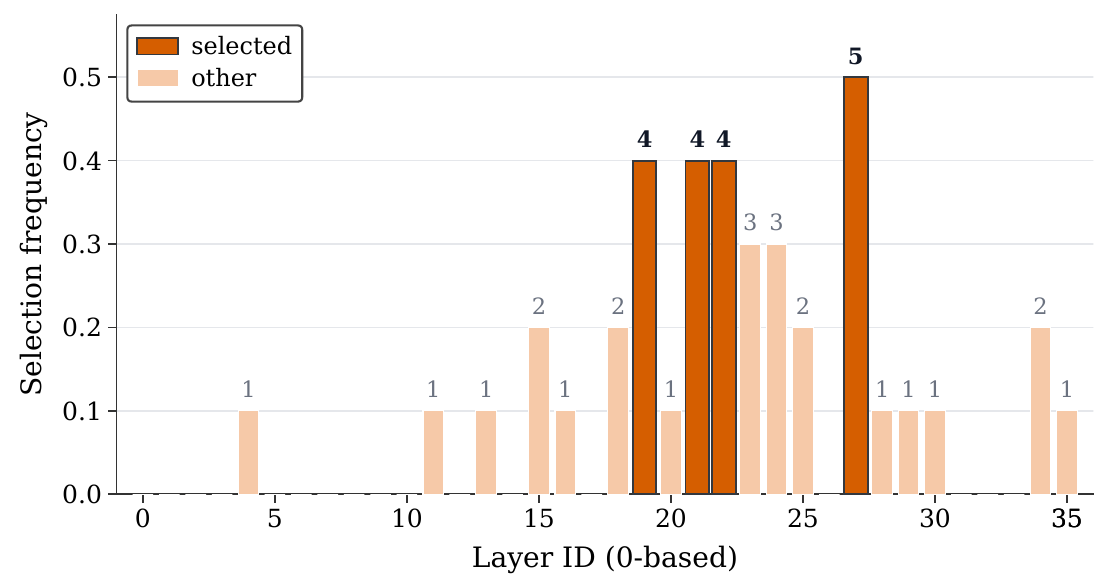}

{\footnotesize (b) Qwen3-4B\par}

\caption{
Layer-selection frequencies over 10 APLS runs on
(a) Qwen3-8B and (b) Qwen3-4B.
Low cross-run agreement indicates non-unique layer subsets.
}
\label{fig:e5_layer_frequency}
\end{figure}

\subsection{Runtime-Policy Ablations}
\label{sec:runtime_ablation}

We isolate the candidate stream and temporal stopping policy. Fig.~\ref{fig:e4_runtime_factors} shows their effects on the MATH-500 accuracy--token frontier.

\paragraph{Candidate stream.}
Fig.~\ref{fig:e4_runtime_factors}(a) compares doubt-only and mixed candidate streams on MATH-500 with Qwen3-8B. By adding ordinary sentence boundaries, the mixed stream introduces more exit opportunities and improves the high-accuracy region of the frontier. This result suggests that self-doubt checkpoints are informative but too sparse to cover all prefixes that already support a reliable answer, motivating broader checkpoint coverage for safe early exit.

\paragraph{Temporal stopping policy.}
Fig.~\ref{fig:e4_runtime_factors}(b) compares temporal stopping policies on MATH-500 with Qwen3-8B. The asymmetric policy exits immediately after a positive self-doubt prediction but requires two consecutive positive sentence predictions. It yields a stronger high-accuracy trade-off than uniform immediate stopping: temporal confirmation reduces premature exits at heterogeneous sentence boundaries while retaining the broader coverage of the mixed candidate stream.

Overall, the strongest runtime frontier is obtained by combining mixed
candidate coverage and asymmetric temporal
stopping. Thus, candidate expansion alone is insufficient: its benefit
depends on how the additional candidates are calibrated and converted
into stopping decisions.

\subsection{Layer-Selection Stability Analysis}
\label{sec:layer_stability}

Fig.~\ref{fig:e5_layer_frequency}, together with
Table~\ref{tab:e5_layer_stability}, shows that independent APLS runs
have low layer overlap and near-zero rank correlations, while their
AUROC variance remains consistently small across both model scales.
Panels (a) and (b) report the layer-selection frequencies for Qwen3-8B
and Qwen3-4B, respectively. The frequency-aggregated subsets are
$[15,19,31,35]$ for Qwen3-8B and $[19,21,22,27]$ for Qwen3-4B,
further indicating that the preferred absolute layer indices are
model dependent.

One plausible explanation is that prefix-sufficiency information is
redundantly distributed across model depth. Because Transformer residual
representations are progressively updated and partially preserve
information from earlier layers, several different layer combinations may
provide comparable evidence for the same sufficiency decision. The
selection objective may therefore admit multiple functionally similar
solutions rather than a single uniquely optimal subset. Accordingly, the
selector should be interpreted as identifying an effective compact
representation, rather than recovering a unique set of mechanistically
critical layers.
\subsection{Efficiency and Resource Analysis}
\label{sec:resource_analysis}

Finally, we examine whether the compact layer subset identified by APLS
translates into practical resource savings. We compare the deployed
APLS compact probe with the dense cross-layer model used during layer
selection. Both architectures are evaluated under the same controlled
training protocol on Qwen3-8B.

  \begin{table}[t]
  \centering
  \begin{tabular}{lcc}
  \toprule
  Metric & Qwen3-8B & Qwen3-4B \\
  \midrule
  Pairwise Jaccard
  & 0.119 $\pm$ 0.093
  & 0.133 $\pm$ 0.147 \\
  Kendall $\tau$
  & -0.005 $\pm$ 0.132
  & 0.011 $\pm$ 0.136 \\
  Spearman $\rho$
  & -0.002 $\pm$ 0.185
  & 0.012 $\pm$ 0.189 \\
  Full-val AUROC $\uparrow$
  & 0.871 $\pm$ 0.004
  & 0.866 $\pm$ 0.003 \\
  Clean AUROC $\uparrow$
  & 0.883 $\pm$ 0.003
  & 0.881 $\pm$ 0.003 \\
  \bottomrule
  \end{tabular}
  \caption{
  Layer-selection stability across 10 fixed-split runs. Pairwise Jaccard and rank correlations are computed over 45 run pairs, and AUROC over individual runs. Low subset agreement but small AUROC variance indicates stable performance despite non-unique layer choices.
  }
  \label{tab:e5_layer_stability}
  \end{table}

\begin{table}[t]
    \centering
    \small
    \setlength{\tabcolsep}{4.5pt}
    \renewcommand{\arraystretch}{1.08}

    \begin{tabular*}{\columnwidth}{
    @{\extracolsep{\fill}}lrrr@{}
    }
    \toprule
    Architecture
    & Params
    & \shortstack{Peak alloc.\\(MiB)}
    & \shortstack{Time\\(s/epoch)} \\
    \midrule
    Dense cross-layer model
    & 11.83M
    & 1348.8
    & 39.87 \\
    APLS compact probe ($K=4$)
    & \textbf{4.24M}
    & \textbf{209.0}
    & \textbf{3.97} \\
    \bottomrule
    \end{tabular*}

    \caption{Efficiency comparison between the dense cross-layer model and
    the compact APLS probe.}
    \label{tab:qwen8b_resource_comparison}
\end{table}

As shown in Table~\ref{tab:qwen8b_resource_comparison}, the APLS
compact probe substantially reduces the parameter count and peak
allocated memory compared with the dense cross-layer model. It also
reduces the per-epoch training time from approximately 40 seconds to
under 4 seconds. These results show that APLS converts compact
probe-layer selection into concrete computational savings.

\section{Conclusion}

We propose BLADE, a boundary-expanded, layer-adaptive framework for dynamic early exit in efficient LLM reasoning. Experiments on five benchmarks and two Qwen3 backbones show that BLADE reduces generated tokens while largely preserving accuracy. Ablations show that combining sentence and self-doubt boundaries uncovers exit opportunities missed by self-doubt-only monitoring, while automatic layer selection yields compact representations. These findings highlight the complementary roles of checkpoint coverage and adaptive layer selection in efficient reasoning.

\bibliography{aaai2027}

\end{document}